# Suction Grasp Region Prediction using Self-supervised Learning for Object Picking in Dense Clutter

Quanquan Shao[*], Jie Hu, Weiming Wang, Yi Fang, Wenhai Liu, Jin Qi, Jin Ma
Shanghai Jiao Tong University
Shanghai, China
e-mail: *sjtudq@qq.com

*Abstract*—This paper focuses on robotic picking tasks in cluttered scenario. Because of the diversity of poses, types of stack and complicated background in bin picking situation, it is much difficult to recognize and estimate their pose before grasping them. Here, this paper combines Resnet with U-net structure, a special framework of Convolution Neural Networks (CNN), to predict picking region without recognition and pose estimation. And it makes robotic picking system learn picking skills from scratch. At the same time, we train the network end to end with online samples. In the end of this paper, several experiments are conducted to demonstrate the performance of our methods.

*Keywords-component; Self-supervised Learning, suction grasp, nueral networks, dense cluter*

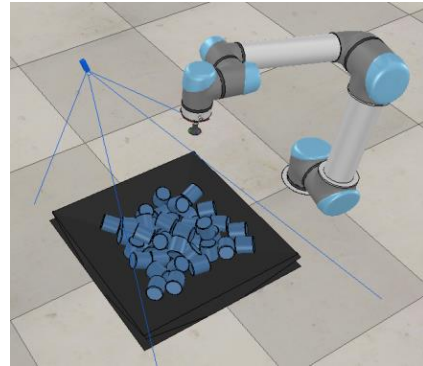

Fig. 1 The suction grasp platform in cluttered environment like bin-picking. Cylindric objects are stacked in an opened box. Robot could pick up objects efficiently with several hours of self-learning.

## I. INTRODUCTION

As one of the basic ways of manipulation technology, robotic grasping technology has a broad application scenario. In industrial situation, it includes automatic loading and unloading, pick-and-place, bin-picking, conveyor tracking, automatic workshop logistics and so on. With the development of electronic commerce, the demand that how robot picking up objects from bins or shelves automatically is growing dramatically. In traditional vision based robotic grasping pipeline, it mainly includes object recognition, pose estimation, grasp point detection and robot execution[1]. Because of the diversity of objects, stack of clutter and complicated background in bin picking situation, it is much difficult to recognize target objects and estimate their pose before grasping them. Object recognition based on 3D-model matching is difficult to deal with various objects and even impossible for novel objects because of the loss of CAD model of target objects[6]. Object self-occlusion and disordered stack also reduce the performance of part 3D matching in robotic picking situation.

Encouraged by success of data-driven methods in many cases, Deep Neural Networks (DNN) were also used to grasp a diverse set of cluttered objects in robotic manipulation research field recently. Deep Convolution Neural Networks (CNN) got a great performance in image classification tasks[5]. And it was widely used in image process and computer vision. Some robotic manipulation researchers transform vision-based object grasping point selection into classification problem and use CNN to decide graspable or not in two-dimensional operation space[10]. Traditional grasp detection hypotheses such as force closure and form closure for generation of grasp candidates were combined with powerful classification performance of CNNs to obtain the best selection of 3D grasping point in cluttered objects[14].

Most of grasping point detection in cluttered scenario are based on parallel-jaw grasping configuration. In this paper, we focus on robotic suction grasps which is also widely applied in robotic manipulation and industrial situation. A novel approach is proposed to detect picking point in disorderly placed known or unknown objects. Images got by RGB-D camera are inputted into a convolution neural network and the output is a probability map which stands for the successful probability of each pixel as the point of suction. Based on this probability map, we could choose a best suck point in cluttered environment like bin-picking, as shown in Fig.1. Furthermore, we train this suction grasp region prediction networks end-to-end with online samples from scratch. It is exciting that robot could get generalized skill of object picking after several hours of suction trials without any help of humans in the learning loop, which could be regarded as a simple reinforcement learning style.

Generally, the main contributions of our method are: a) We introduce a novel method of suction point selection based on a combination of Resnet and U-net framework in bin-picking situation which makes the selection of suction point as a suction region prediction problem. b) We train this network end to end with self-supervised learning. As a result, we could get an effective robotic picking system after a few hours of self-learning without human help.

The remaining of this paper is structured as follows: In section 2, some related works about object grasping, grasping point detection based on deep neural networks, Resnet and U-net framework are discussed and then we elaborately present our region suction point detection method with a deep neural network and its online train method with a self-learning way in section 3. Several experiments are conducted to demonstrate this pipeline and detailed procedure with their results are introduced in section 4. Finally, the conclusion and future works are given in section 5.

## II. RELATED WORK

Object grasping is widely explored in the robotic research area, but there are still many problems to be solved because of the variety of environment and the uncertainty of perception and execution. Early studies mainly research the grasp plan from perspective of force-closure and form-closure which need get the shape and pose of the object. Howard analyzed the stability of grasped object in a situation that force not closed[4]. Zhu and Ding proposed an iterative algorithm computing the form and force closure for the grasp of 3D object[15]. It is unrealistic to obtain 3D models of all objects. Data-driven methods are used with a variety of different forms to get more generalized grasp skills. These methods predict the grasp configuration directly during test time without the estimation of the pose and shape of target objects[2]. With the tremendous development of deep learning, data-driven grasp detection are mostly combined with deep neural networks in recent research[7][10][14].

Lenz applied the feature learning character of deep neural network to detect robotic grasps with RGB-D perception[7]. Redmon expanded the method of Lenz and used a CNN to robotic grasp detection with a faster speed[12]. They were mainly studied in a separate situation, not in cluttered scenario. Pinto studied grasp detection based on deep learning from another perspective. He proposed one self-supervision grasp detection methods getting training datasets online[10], which was low efficient and needed 700 hours. Levine also used an on-line framework for 2D grasping detection without hand-eye calibration, which shows the end-to-end powerful of deep neural networks[8]. However, it needed 6~14 robots for several months to collect training data. As we said before, Pas used partially occluded RGB-D or point clouds to generate lots of grasp candidates and evaluated these candidates with a convolution neural network[14]. The method was very effective in object- agnostic cluttered scenario. Parallel-jaw and multi-finger grasp were exploited broadly. Mahler proposed a Grasp Quality Convolution Neural Network (GQ-CNN) to estimate the quality of suction grasp based on point clouds, which used 1500 3D object models to train the network. Zeng applied fully convolutional networks to suction grasp detection and took the first place in Amazon Robotics Challenge[16]. Zeng also combined pushing and grasping with self-supervised with two-finger gripper[16][17], however, it needed reprojection before inputting prediction network. Inspired by above research, we directly use the perception information of RGB-D camera to input prediction network, namely a framework combined Resnet and U-net structure and train the network with a self-learning style. Resnet was proposed by Kaiming He, and he added identity shortcut connections with different layers and got a higher classification accuracy in ImageNet in 2015[18]. U-net was first used for biomedical image segmentation[13], which combined pooling operation and up-sampling operation and could also get good performance with few images. Based on its relative high training efficiency, it was also used to 3D segmentation in biomedical information process[2].

## III. METHODOLOGY

Our aim is to construct a self-supervised online learning robotic bin-picking system. The main framework of this system is shown in Fig. 2. Colored cylindric objects are stacked in an opened box. And robot could pick up objects efficiently with several hours of self-learning. Perception part of this robot system is RGB-D camera, which could output RGB images and depth maps. With these visual information, suction grasp region prediction network gives success suction probability maps. It is no doubt that objects could be sucked in regions in the image coordinate system. Suction point is chosen based with the maximum value pixel. The point with most success value will be executed to check the performance of this prediction network. Finally, this region prediction network is trained based on the result of executed robot action. Most important parts of this self-learning robotic picking system are the suction region prediction network and the framework of self-supervised learning with this network.

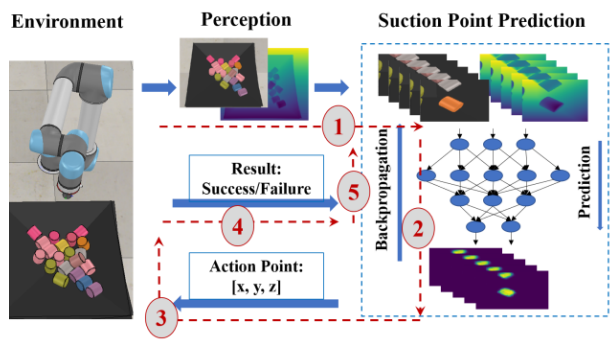

Fig. 2 The structure of self-learning robotic picking system. Robot learns to suck cylindric objects in a cluttered bin. First, images and depth maps obtained by RGB-D camera are split into 168×168 image patches. And then these patches are inputted into region prediction networks and a success suction probability map with 56×56 is given. Third, suction point is chosen based on this map and transformed into the world frame. The robot executes picking action with this suction point subsequently. At last, Prediction network is trained with executed result.

### A. Suction Region Prediction Network

The Resnet and U-net framework had been successfully used in image classification and region detection with images. In this paper, we use Resnet pretrained with ImageNet datasets to obtain different scale feature maps and U-net framework to predict suction region. In the training phase, parameters of Resnet parts are fixed and we only adjust parameters of U-net framework parts. Detailed framework of region prediction network is shown in Fig. 3.

To reduce numbers of parameters, the size of inputs of this network is chosen as 168 ×168. Firstly, these image patches are resized to 224×224, default input size of Resnet. Three channels of depth stream are the same with depth maps. Outputs are maps of two channels with size $56 \times 56$. And the first channel means success probability maps, the other is counterpart based on bi-level classification. Each pixel of this probability map represents the success probability when robot trying to pick objects at central of 3×3 patch images of input images correspondingly. That means suction point has step of three pixels in image coordinate frame. As shown in Fig. 3, Blue box is RGB stream and pink box stands for depth stream. These two networks are Resnet50 with same parameters, which are downloaded from official github website of tensorflow[19]. Parameters of Resnet50 are fixed in learning phase. Green boxes in Fig. 3 are convolution layers with U-net structure. Parameters of U-net are adjusted to obtain a valid suction region map.

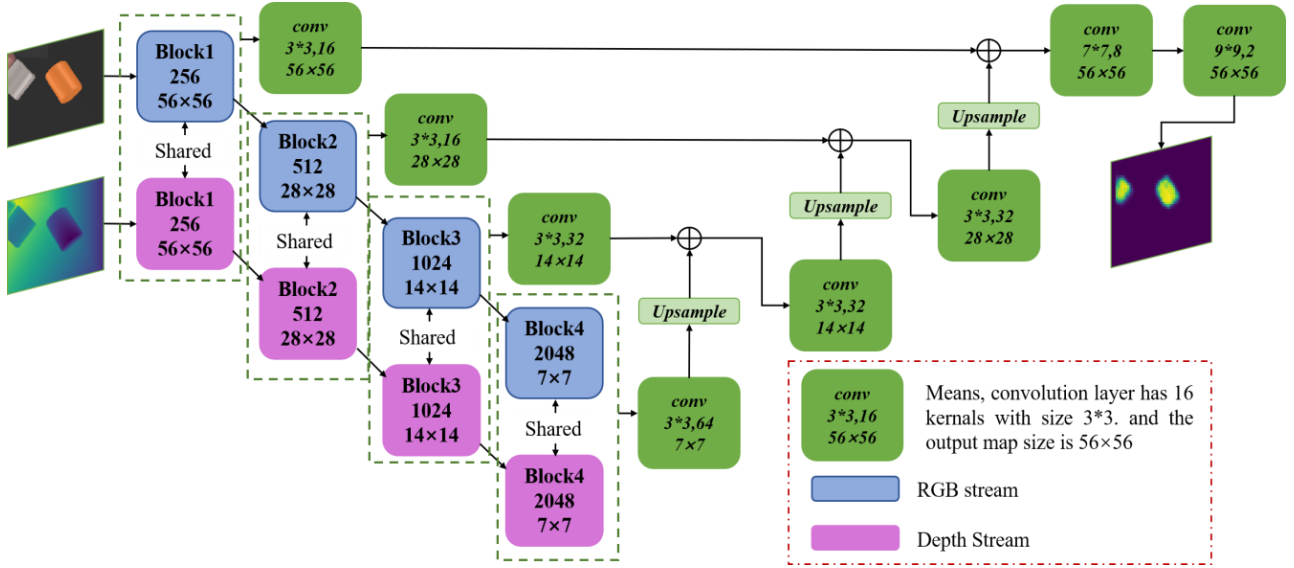

Fig. 3 Structure of suction grasp affordance region prediction network. Blue boxes mean RGB image stream and pink boxes mean depth map stream. Numbers and sizes of output maps of each block are shown in box. These two Resnet streams have the same parameters in corresponding scale. And they are trained with ImageNet and fixed in this application. Green boxes are convolution layers with U-net structure. "Upsample" boxes mean twofold resize without parameters to be trained. Parameters of these convolution layers will be adjusted in training phase.

*B. Executed Suction Point*

After getting this suction region map, we could choose the best suction point in image frame. Three-dimensional coordinates of the selected point are obtained by combining index of pixel selected and depth value with projection matrix. Specific relationship is shown below.

$$\begin{bmatrix} x \\ y \\ z \end{bmatrix} = z\Sigma^{-1} \begin{bmatrix} u \\ v \\ 1 \end{bmatrix} \quad (1)$$

where $\Sigma$ is projection matrix of the RGB sensor, which is the intrinsic parameters of camera, $[x, y, z]^T$ are coordinates of point in sensor local frame, $z$ is the value in depth map and $[u, v, 1]^T$ stands for generalized coordinates of each point in depth map. And then executed suction point in robotic frame could be obtained after a simple transformation from local camera coordination to world coordination.

$$\begin{bmatrix} x \\ y \\ z \\ 1 \end{bmatrix}_w = T_{c2w} \begin{bmatrix} x \\ y \\ z \\ 1 \end{bmatrix}_{loc} \quad (2)$$

where $T_{c2w}$ is transformation matrix from camera frame to robotic world frame. $[x, y, z, 1]_w^T$ and $[x, y, z, 1]_{loc}^T$ are generalized coordinates of suction point in world frame and camera frame respectively. We only consider vertical suction in this scenario. Therefore, robot could execute picking actions only using position coordinates without attitude information.

*C. Framework of Self-supervised Learning*

As mentioned above, in the prediction network, Resnet is trained with ImageNet datasets while U-net is not be trained. Trained Resnet could give different scale feature maps. We use U-net to combine these multi-scale feature maps to obtain suction region maps. Parameters of U-net would be trained in an online form. They are initialized randomly with gaussian distribution. And then we use this initialized network to predict suction region maps. The most probable point or a random selected point is transmitted to robot platform and robot tries to pick object at selected suction point in world frame. If robot sucked objects successfully, the corresponding pixel in predicted region map should be set one otherwise it should be zero. We could also treat this suction region prediction problem as a binary classification problem for each pixel in prediction maps. Ideal values should be one or zero based on results of robotic actions and there are some errors between ideal values and the probability predicted by region prediction network. Therefore, the network could be trained online by propagating back the errors. To clarify, we only consider errors of the selected point in probability map. The loss is formulated below.

$$Loss = \frac{1}{N}\sum -y\log(\tilde{y}) - (1-y)\log(1-\tilde{y}) \quad (3)$$

where N is the number of selected pixels, y is value of selected point in prediction map, $\tilde{y}$ is ideal value of selected suction point. $\tilde{y}$ is one or zero based on result of robotic action. Actually, loss is mean of cross entropy of selected point. The workflow of self-supervised learning algorithm of this robotic picking system is shown in Algorithm 1.

It is important to balance exploration and exploitation in reinforcement learning methods. Here, we choose suction point randomly with a varying probability in training phase. The probabilistic value could be obtained with equation 4.

$$p_{ep} = 0.5 \times 5^{(-\frac{n_l}{N_{ep}})} \quad (4)$$

where $p_{ep}$ is the exploration probability with random choosing suction point in image frame, $n_l$ is the training step in self-learning phase, $N_{ep}$ is hyper-parameter to control the decay rate of exploration probability based on number of trials.

**Algorithm 1** self-supervised learning picking
---
**Input:** max_step, hyper-parameter
**Output:** parameters of trained network
1: initialize network and max step
2: **loop:**
3:    (I, D)=get_img    //get visual information
4:    ($I_n, D_n$)=split(I, D)    //split images
5:    M=network($I_n, D_n$)    //region prediction
6:    //7-11 select point with exploration and exploitation
7:    prob=explore(step)
8:    **if** rand(0,1)>prob :
9:       $P_{img}, I_k, D_k$=argmax(M)    //argmax select
10:    **else:**
11:       $P_{img}, I_k, D_k$=random()    //random select
12:    $P_{world}$=trans($P_{img}$)    //get world pose
13:    R=exc($P_{world}$)    // robot executed, return result
14:    $L_k$=label( R )    // label with executing result
15:    train($L_k, I_k, D_k$)    // train network

## IV. EXPERIMENT AND RESULT

In order to validate the feasibility of proposed algorithm for robotic picking manipulation, a series of experiments in V-REP physical simulation environment are conducted in this section. The robotic suction grasp platform in cluttered bin-picking environment is shown in Fig. 1. Cylindric objects are stacked in an opened box. Robot tries to learn to pick up objects efficiently by constantly trying. Simulate UR5 robot equipped with vacuum cup is chosen. The simulate RGB-D camera gives images and depth maps with size $640 \times 480$. The suction region prediction network is structured with Tensorflow1.7, a machine learning system published by Google. And the hardware is a notebook with a 2.6GHz Intel Core i7-6700HQ CPU and a NVIDIA GTX 965 GPU. And the operation system of this notebook is Ubuntu16.04 LTS. Experiments are mainly classified into two categories, self-learning phase and testing phase.

### A. Self-learning Phase

Some experiments are conducted to evaluate effectiveness of this self-learning method. We make robot to execute 2000 attempts, namely, max step in algorithm 1 is 2000. $N_{ep}$ is 1000 in equation 4. The prediction network is trained with gradient optimizer with momentum and relative hyper-parameters are followed: learning rates of $5 \times 10^{-5}$, momentum of 0.9 and regularization coefficient of $1 \times 10^{-4}$. Four types of objects stacking are chosen randomly in online training phase, which are shown in Fig. 4. They are totally twenty cylindric objects.

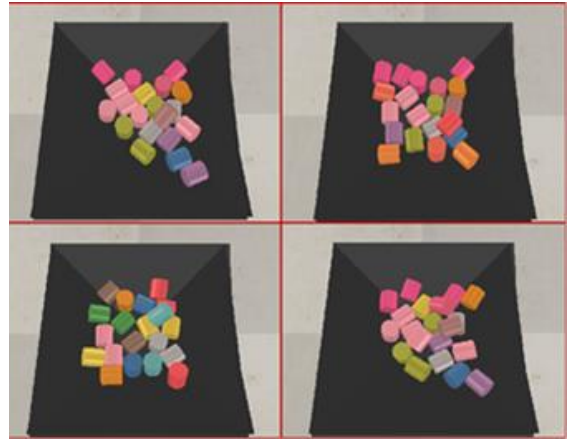

Fig. 4 Four types of object stacking in the training phase. We randomly choose one scene when the bin is empty During the online learning phase.

The evaluation index is success rate per nearest hundred trials, which could be formulated in equation 5.

$$p_{scs} = \frac{n_{scs}}{100} \quad (5)$$

Where $n_{scs}$ is the number of successful trials in latest one hundred attempts. The learning curve is shown in Fig. 5. We could find that this robotic suction system learns to pick objects efficiently with two thousand attempts. In this picture, we also compare results with different inputs, namely, RGB-D inputs and double RGB inputs. The blue curve is result with RGBD inputs as introduced before. The pink curve is result of inputs with Double RGB images with the same structure network, which means the depth stream is also inputted with RGB images. We find that RGB-D information could give high success rate, as shown in Fig. 5.

It should be noted that experiments in Fig. 5 are conducted without data augmentation in network training phase to demonstrate the effectiveness of our algorithm. Next, we introduce data augmentation (DA). The online labeled images are rotated, flipped and mirrored respectively, and they are all inputted into the network for training at the same time. Results are shown in Fig. 6. The pink curve is result with data augmentation while the blue curve is result without data augmentation. Although, for depth maps, rotating or flipping process will break the information of the index of pixel and its value. We could find that data augmentation also gives a weakly better result. That because data augmentation increases diversity of the samples.

We further study the performance of this self-learning framework with network pretraining. We label thirty images manually, which is not very precise and use these coarsely labeled images to pretrain the network before online learning. Result is shown in Fig. 7. It indicates that network pretrained with labeled samples could get a better initial result but there is no help for the end result. However, pretraining process could reduce the number of trials while introducing bias of human labeled images. As shown below, network pretrained with labeled samples has a weak generalization ability.

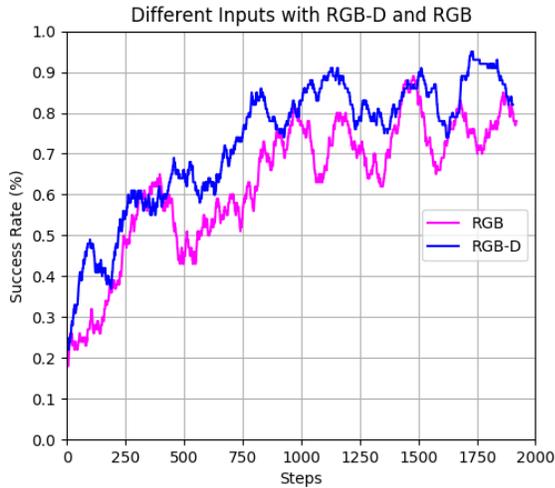

Fig. 5 Results with different inputs. Pink curve is result with double RGB images. Blue curve is result with RGB images and Depth maps. RGBD information could obtain better result.

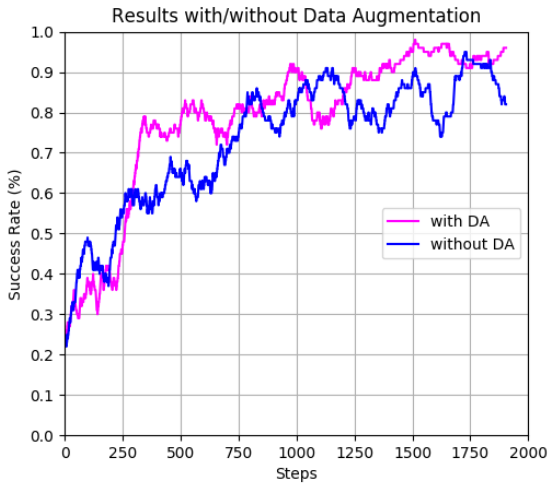

Fig. 6 Results with or without data augmentation. Pink curve is the result with data augmentation. Blue curve is the result without data augmentation. Data augmentation gives a weakly better result.

*B. Validation of Generalization*

To validate the generalization of this robotic suction grasp affordance region prediction network, some experiments are conducted with old and new scenes. Three testing scenes are shown in Fig. 8. There are 20 colored cylindric objects in testing one which is not trained in learning phase and it has the same number of objects with scenes in training phase. Testing two and testing three are all 40 objects which are twice the number of objects in training phase. That is no doubt that they are more complex. Testing two have different colored objects while testing three have all the blue objects, shown in Fig. 7.

There are twenty cylindric objects in testing 1, and the robot will repeatedly clean the box ten times. In other words, two hundred successful suction will be executed. For testing 2 and testing 3, there are forty cylindric objects and we let robot repeatedly pick all objects in the bin five times.

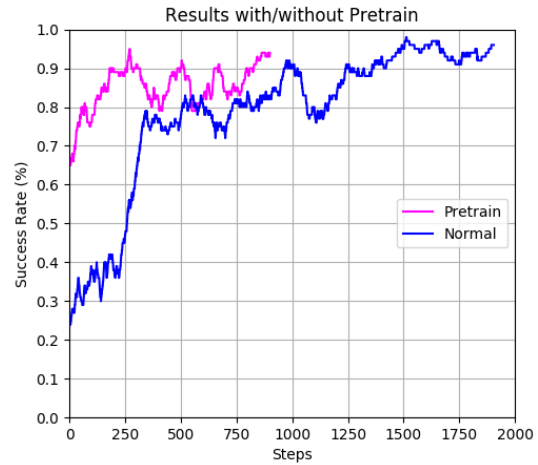

Fig. 7 Results with or without pretrain. Pink curve is the result of pretrain and self-learning with 1000 trials. Blue curve is result without pretrain.

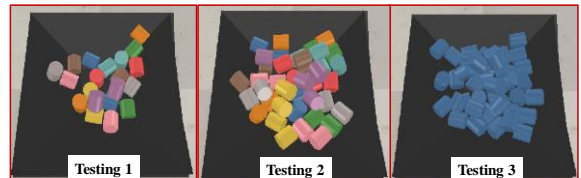

Fig. 7 Three testing cases in the testing phase. Testing 1 have 20 colored cylindric objects randomly stacking in a bin. Testing 2 have 40 colored cylindric objects. And testing 3 have 40 cylindric objects, which are all blue.

Some experiments are executed to compare the networks trained with or without pretrain, introduced in Fig. 7. We make robot clean up objects ten times in testing 1 and scene used in training phase. Result is shown in Table I. 217 is the total number of trails to pick up all objects 10 times and 92.17% is the average success rate. Network pretrained with labeled samples has a weak performance of generalization which drops from 92.17% to 88.11% in training scene and new scene. While network learning from scratch could get high success rate in all these two scenes, which is 97.56% and 96.62% respectively.

TABLE I.  RESULT WITH/WITHOUT PRETRAIN

| Items | Result of different scene | |
|---|---|---|
| | *Scene used in training phase* | *Testing 1* |
| **Network with Pretrain** | 217/92.17% | 227/88.11% |
| **Network without Pretrain** | 205/97.56% | 207/96.62% |

Further, we conduct some experiments with network learned from scratch in testing scene 1, 2 and 3 to validate its generalization ability. we perform 10 or 5 times for different tests. Details are shown in Table II and success rate are 96.62%, 79.05% and 85.11% respectively.

From testing 1, we could find the suction grasp region prediction network is also efficiently in unseen scenario, which has different background and different various pose for each robotic suction. That means it has a relative strong generalization ability. Compared with testing 1, testing 2 has forty colored cylindric objects and there are more possible postures and more complex surroundings. It has a bit weak performance with success rate 79.05%. For the last scene, it also has forty cylindric objects but they are all the same color. Success rate is 85.11%, which is better than testing 2 and worse than testing 1. Compared with testing 1, testing 3 has 40 objects and the reasons of the weakness of success rate are similar with testing 2. Compared with testing 2, all objects are the same color in testing 3 and this difference eliminates the preference of this suction grasp region prediction network for color and focuses on the posture and surroundings. It is the reason why prediction network could get a higher performance in testing 3. We believe that it could have better result in testing 2 and testing 3, if we trained the network with 40 objects. In conclusion, this self-learned network has a great generalization ability.

TABLE II. TESTING RESULT WITHOUT PRETRAIIN

| Items | Result of different scene | | |
|---|---|---|---|
| | *Testing 1* | *Testing 2* | *Testing 3* |
| Total Objects | 20 | 40 | 40 |
| Repeat Times | 10 | 5 | 5 |
| Success Suction | 200 | 200 | 200 |
| Failure Suction | 7 | 53 | 35 |
| Total Suction | 207 | 253 | 235 |
| Success Rate | **0.9662** | **0.7905** | **0.8511** |

## V. CONCLUSION

This paper proposes a new method which uses a region detection convolution framework to detect suction point in dense clutter and an effective self-learning framework for the train of the network. This method predicts suction grasp region for object picking without recognition and pose estimation which is very appropriate in object-agnostic scenario. Experimental results demonstrate the validity of our methods.

In future work, we prepare to combine object recognition and grasp point detection with the same network which means we pick one object with the high probability of target object and the high probability of graspability in weak recognition conditions, just like dense clutter. And robot could re-estimate it after picked out just like the circumstance that human fetch objects with the most probability of target objects and then judge whether it is the desired object in a weak recognition condition. We also explore ways to enhance the learning speed and the robust of this method in more complex scenes.

ACKNOWLEDGMENT

This research is supported by Special Program for Innovation Method of the Ministry of Science and Technology.